\documentclass[runningheads]{llncs}
\usepackage{graphicx}
\usepackage{booktabs}
\usepackage{hyperref}
\usepackage{comment}
\usepackage{multirow}
\usepackage{array}
\usepackage{subcaption}
\usepackage{amsmath}

\newcommand{\DR}{DR}

\begin{document}
\title{Targeted Visualization of the Backbone of Encoder LLMs\thanks{This project is funded by the European Union through Project LEMUR.}} 

\author{Isaac Roberts\inst{1} \and
Alexander Schulz\inst{1}\and
Luca Hermes\inst{1}\and
Barbara Hammer\inst{1}}
\authorrunning{I. Roberts et al.}

\institute{Machine Learning Group, Bielefeld University, D-33619 Bielefeld, Germany
\email{\{iroberts,aschulz,lhermes,bhammer\}@techfak.uni-bielefeld.de}}
\maketitle              
\begin{abstract}
Attention based Large Language Models (LLMs) are the state-of-the-art in natural language processing (NLP). The two most common architectures are encoders such as BERT, and decoders like the GPT models.  Despite the success of encoder models, on which we focus in this work, they also bear several risks, including issues with bias or their susceptibility for adversarial attacks, signifying the necessity for explainable AI to detect such issues.
While there does exist various local explainability methods focusing on the prediction of single inputs, global methods based on dimensionality reduction for classification inspection,
which have emerged in other domains and that go further than just using t-SNE in the embedding space, are not widely spread in NLP.

To reduce this gap, we investigate the application of DeepView, a method for visualizing a part of the decision function together with a data set in two dimensions, to the NLP domain. While in previous work, DeepView has been used to inspect deep image classification models, we demonstrate how to apply it to BERT-based NLP classifiers and investigate its usability in this domain, including settings with adversarially perturbed input samples and pre-trained, fine-tuned, and multi-task models.

\end{abstract}
\section{Introduction}

Since the development of the attention architecture \cite{attention}, large pre-trained language models such as encoders like BERT \cite{bert} and decoders like GPT \cite{GPT} have become a dominant technology in many tasks of NLP, which is under continuous further development \cite{GPT4,Llama2} and includes extensions to e.g.\ chat-bots. 
However, even when focusing only on encoder models, issues and risks of such models do exist when it comes to downstream tasks \cite{llmethical}. These include robustness, for instance due to adversarial examples \cite{adversaryBert,ribeiro-etal-2020-beyond}, or observing social biases  
\cite{seat,weat,biasbios}. 
Such risks demonstrate the need of explainable AI (xAI) methods in order to help detect, better understand and mitigate them. 

Explanations for encoder models can be categorized into local explanations, where the goal is to explain the behaviour of the model for a specific given input instance, and global explanations, which aim to provide insights into the workings of the model on a broader level. Regarding local explanations in encoder-based LLMs, they can be grouped into feature attribution approaches, attention-based explanations, example and natural-language-based methods \cite{explainingLLMs}.
Global explanations can be grouped into probing-based, neuron activation, concept-based, mechanistic \cite{explainingLLMs} and also dimensionality-reduction-based \cite{bertvis}. The latter, however, in NLP often boil down to a simple application of nonlinear dimensionality reduction (\DR).
In other domains, however, tailored \DR\ approaches for explanation have emerged to provide useful inspections of classifiers \cite{deepview,spray,dvi}.

In this work, we focus on dimensionality reduction-based approaches and investigate the application of DeepView \cite{deepview}, a recently proposed tool to visually inspect the decision function of a high-dimensional classifier with regard to a given data set. It extends dimensionality reduction methods with a discriminative component and with the projection of a part of the decision function.
Since DeepView has only been applied to the image domain, we analyse how to utilize it for encoder-only LLMs and showcase its usefulness in practical applications.

More precisely, our contributions are (i) a procedure to apply DeepView to different settings of encoder based NLP models including frozen, fine-tuned and multi-task fine-tuned BERT encoders, (ii) application examples where DeepView supports the finding of interesting information, such as adversarially perturbed and atypical samples among normal data, the latent space organization helping to detect synergy between different tasks and examples of similarity between classification strategies of different models.
Our code is available online\footnote{\url{https://github.com/LucaHermes/DeepView}}.

In the remainder of the paper, we first recap some foundations in section \ref{sec:bg}, describe our modifications of DeepView in \ref{sec:deepview_mod}, provide details regarding the trained encoder models in \ref{sec:training_details}, present our experiments in \ref{sec:exper} and conclude in \ref{sec:conclusion}.

\section{Background} \label{sec:bg}
This section recaps the \DR\ literature, that we build upon, including the classifier visualization tool DeepView \cite{deepview} and methods to evaluate \DR\ mappings.

\subsection{DeepView}

DeepView \cite{deepview} is a framework to visualize a part of the decision function of a deep neural network classifier together with a data subset in two dimensions. It consists of four steps: (i) project the data down using a discriminative \DR\ mapping, (ii) sample a regular grid in the projection space and project it to the original data space, (iii) apply the classifier to the projected samples to obtain the class label and the certainty estimate and (iv) visualize these in the background of a scatter plot to obtain an approximation of the decision function. 

The first step is the most important one as it selects the subspace for visualization. For this purpose, DeepView utilizes a discriminative (sometimes also referred to as supervised) variant of UMAP \cite{umap}, which uses regular UMAP together with a discriminative distance metric that emphasizes directions in the data space, where classifier predictions change. 
More formally, let $f$ be the classifier outputting a probability distribution over the classes and $d_{JS}$ be the Jensen-Shannon metric $d_{JS}(p,q) := \sqrt{(D_{KL}(p \Vert m)+D_{KL}(q \Vert m))/2}, m = (p+q)/2$ based on the Kullback-Leibler divergence $D_{KL}$. DeepView utilizes
\begin{align}
	d(x,y) := \sum_{i = 1}^n (1-\lambda) d_{JS}\left(f\left(  p_{i-1}  \right),f\left( p_i \right)\right)+\lambda d_S\left( p_{i-1},p_i \right) \label{eq:dist}
\end{align}
as the discriminative distance metric between two points from the original input space $x$ and $y$, where the sum is over $n$ equidistant points $p_i = \left( 1- \frac{i}{n}\right)x + \frac{i}{n} y$ on a straight line from $x$ to $y$ approximating an arc-length metric, $d_S$ is an unsupervised default metric on the data space acting as a regularization, e.g.\ the euclidean metric, and $\lambda\in [0,1]$ determines the balance between the discriminative and unsupervised metric.

After the \DR\ has been used to select a subspace of the input for visualization, an inverse mapping is required to project the corresponding part of the decision function. A suitable mapping when using UMAP for \DR\ has been shown in \cite{deepview} to be a Radial Basis Function network.

In addition to numerical evaluation measures (see \ref{sec:qknn}), we also make use of visual cues in the DeepView images to help users judge parts of the plot. To this end, a larger circle surrounding a point in the scatter plot encodes that the model's predicted value of the input is different than the background in that area, while the true label of each input is indicated by the color of the point.

\subsection{Evaluation in Dimensionality Reduction} \label{sec:dr_eval}

Common methods for evaluating the quality of an unsupervised dimensionality reduction mapping are summarized in \cite{pyDRMetrics}. Popular strategies are based on the neighborhood preservation of the \DR. Here, we utilize the following measures, which are applied to two representations of the same data points, usually the original data space and the embedding: \\
$\mathbf{Q_{NN}(k)}$: How many points are the same among the k nearest neighbors between the two data representations. 
$\mathbf{LCMC(k)}$: The baseline value, which corresponds to the diagonal of the $Q_{NN}(k)$ curve, is removed from $Q_{NN}(k)$.
$\mathbf{K_{max}}$: Corresponds to the maximum value of $LCMC(k)$. 
$\mathbf{AUC}(Q_{NN}(k))$ is the area under the $Q_{NN}(k)$ curve.
$\mathbf{Q_{local}}$ is the area the $Q_{NN}(k)$ curve up to $K_{max}$ with normalization, highlighting local neighborhood preservation.

However, these rely classically on an unsupervised metric in the input space and hence are not useful to evaluate discriminative projections. We will utilize them in a slightly different way, by applying to two embedding spaces and thereby comparing the similarity of these.

\noindent \textbf{DeepView Evaluation} \label{sec:qknn} To assess the quality of the DeepView \DR\ projection, 
we evaluate its fidelity to the model's decision function. This is determined by the extent to which the projected data reflect similar classification outcomes for proximal points in the embedding space, providing insights into how a model organizes its embedding space by class. Following \cite{deepview}, we gauge the fidelity using $Q_{kNN}$, which calculates the leave-one-out error of a k-nearest neighbor (kNN) classifier with $k=5$ using the model's predicted labels. 
We also evaluate the quality of the inverse projection by reporting the agreement between the prediction labels of data points and the background of the scatter plot as $Q_{data}$.

\section{DeepView Modification for NLP} \label{sec:deepview_mod}
Previous applications of DeepView focused on the image classification domain. In order to adapt it for use in the NLP domain, we propose to make the following adjustments: (i) using the BERT embedding as the input space for DeepView and (ii) employing the cosine distance for regularization. We also provide useful pipelines and demonstration examples in section \ref{sec:exper}.

To apply DeepView, an input space with a distance measure and the possibility to compute interpolations between data points is required. For this purpose, we leverage the encoding of the BERT model \cite{bert} using the "CLS" token to create a single vector per input sequence and employ this as the data representation. 

Further, DeepView utilizes a discriminative distance measure and regularizes it with $d_S$, an unsupervised one. While the discriminative distance depends on the classifier in question, the cosine distance is the most natural choice for regularization in the present setting. 
With this framework, we will investigate how dominant the discriminative information is in the embedding space of different models, among other experiments.

\section{BERT Training Set-up} \label{sec:training_details}

\begin{table}[t]
\centering
\caption{
Classification accuracies for all models. Additionally Matthew's Correlation for COLA and F1 score for MRPC due to class imbalance.
}\label{tab3}
\begin{tabular}{l c p{0.12\textwidth}<{\centering} p{0.12\textwidth}<{\centering}cccc}
\toprule
\textbf{Setting} &  \textbf{SST2} & \textbf{COLA} & \textbf{MRPC} & \textbf{QQP} & \textbf{QNLI} & \textbf{MNLI} & \textbf{RTE} \\
\midrule
Pre-trained BERT & 0.80 & 0.74/0.37 & 0.68/0.41 & 0.70 & 0.70 & 0.47 & 0.56  \\
Fine-tuned BERT & 0.92 & 0.81/0.52 & 0.84/0.89 & 0.88 & 0.91 & 0.84 & 0.68 \\
Multi-Task BERT & 0.91 & 0.80/0.49 & 0.84/0.88 & 0.90 & 0.90 & 0.83 & 0.76  \\
\bottomrule
\end{tabular}
\end{table}

This section summarizes the technical details of the NLP classification models that we inspect later.

\noindent \textbf{Datasets} The Glue dataset \cite{wang2019glue} comprises a collection of datasets used for various NLP tasks that are phrased as classification problems. For the Single Sentence Sentiment Analysis task, we utilize the Stanford Sentiment Treebank (SST2) data set, for Single Sentence Grammar Acceptability task the Corpus of Linguistic Acceptability (COLA) set, for Sentence Similarity the Microsoft Research Paraphrase Corpus (MRPC) and Quora Question Pairs (QQP) sets, and for Natural Language Inference the Stanford Question Answering Dataset (QNLI), Multi-Genre Natural Language Inference (MNLI) and Recognizing Textual Entailment (RTE) datasets.

\noindent \textbf{Model Architectures} In our study, we utilize three standard types of training to investigate differences among them using DeepView. We always employ standard pre-trained BERT.
\textbf{Pre-trained} (PT) BERT refers to a frozen encoder model and training only the additional classification head. It consists of three fully-connected layers with sizes $768-512-256$ for all data sets, except for MNLI where one layer performed best.
For \textbf{fine-tuned} (FT) BERT the encoder as well as the classification head were fine-tuned to one task, and as such, the classification head consists of a single linear layer.
\textbf{Multi-Task} (MT) BERT refers to fine-tuning the whole model as well, but with an individual classification head for each task and using all tasks simultaneously and shuffled for training.

\noindent \textbf{Training Hyperparameters} We detail the performance of the resulting models on the data sets in \ref{tab3}, where we always report the accuracy and for the data sets with class imbalance (COLA and MRPC) additionally a metric that is invariant to class imbalance.
The used hyperparameters are desciribed in the following:
All models use the "bert-based-uncased" tokenization and model engine, with Hugging Face's default optimizer Adam Weighted Decay, using a batch size of 16 and learning rate of $2e{-05}$. The multi-task model is trained for 3 epochs, the fine-tuned ones for 5 on the smaller datasets (MRPC, RTE, COLA) and 3 epochs for the larger ones (MNLI, QNLI, QQP, SST2). The pre-trained models are trained similarly, except that COLA was trained for 10 epochs and QNLI for 9 epochs for better performance.

\section{Experiments} \label{sec:exper}

\begin{table}[t]
\centering
\caption{Mean and standard deviations of $Q_{kNN}$ and $Q_{data}$ percent errors with different weights of discriminative and unsupervised metrics. PT values averaged over GLUE Datasets where the classifier performed better than random.}\label{tab4}
\begin{tabular}{c c p{0.12\textwidth}<{\centering} p{0.12\textwidth}<{\centering} p{0.12\textwidth}<{\centering} p{0.12\textwidth}<{\centering} p{0.12\textwidth}<{\centering} p{0.12\textwidth}<{\centering}}
\toprule
& \textbf{Setting} & \textbf{$\lambda$ = 1} & \textbf{$\lambda$ = 0.8} & \textbf{$\lambda$ = 0.6} & \textbf{$\lambda$ = 0.4} & \textbf{$\lambda$ = 0.2} & \textbf{$\lambda$ = 0.0} \\
        \midrule

    \parbox[t]{2mm}{\multirow{3}{*}{\rotatebox[origin=c]{90}{$Q_{kNN}$}}} & 
    PT & $27 \pm 12$  & $2 \pm 4$ & $2 \pm 3$ & $1 \pm 2$ & $2 \pm 2$ & $1 \pm 2$ \\
    & FT & $2 \pm 2$  & $2 \pm 2$  & $1 \pm 1$ & $1 \pm 1$  & $1 \pm 1$ & $1 \pm 1$   \\
    & MT & $1 \pm 1$  & $1 \pm 1$ & $1 \pm 1$ & $ 1 \pm 1$ & $1 \pm 1$ & $1 \pm 1$ \\
    \parbox[t]{2mm}{\multirow{3}{*}{\rotatebox[origin=c]{90}{$Q_{data}$}}} &
    PT & $31 \pm 14$  & $3 \pm 6$ & $3 \pm 4$ & $2 \pm 3$ & $0.3 \pm 0.6$ & $1 \pm 1$ \\
    & FT & $1 \pm 1$  & $1 \pm 1$  & $1 \pm 1$ & $0.3 \pm 1$  & $0.3 \pm 1$ & $0.3 \pm 1$   \\
    & MT & $1 \pm 1$  & $1 \pm 1$ & $1 \pm 1$ & $1 \pm 1$ & $0.3 \pm 1$ & $0 \pm 0$ \\
\bottomrule
\end{tabular}
\end{table}

In this section, we delve into BERT's embedding space and classification function using DeepView across various settings, focusing on the tasks given by the Glue Dataset. Section \ref{sec:exper_disc} begins by evaluating the use of  discriminative information in visualization of BERT-based classification models.

We demonstrate DeepView's application in current NLP settings, showcasing its utility in identifying adversarials (Sec. \ref{sec:exper_adv}), understanding multi-task model organization (Sec. \ref{sec:exper_multi}), and comparing classification strategies across models (Sec. \ref{sec:exper_neigh}).

\subsection{Information Provided by Discriminiative Distances}\label{sec:exper_disc}

\begin{figure}[tb]
    \centering
    \begin{subfigure}[b]{0.25\textwidth}
        \centering
        \includegraphics[width=\textwidth,trim= 0 80 0 25, clip=true]{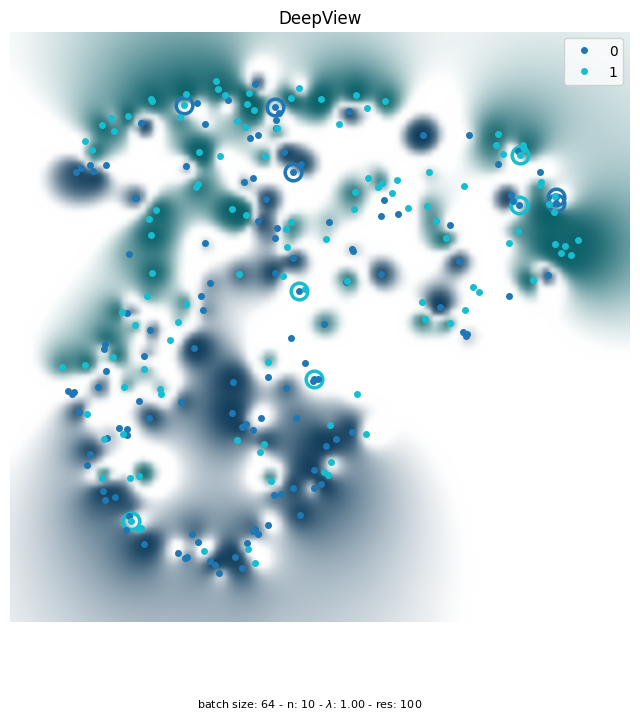}
        \caption{$\lambda=1$}
        \label{subfig:lambda1}
    \end{subfigure}
    \begin{subfigure}[b]{0.35\textwidth}
        \centering
        \includegraphics[width=\textwidth, trim= 0 175 0 25, clip=true, angle=0]{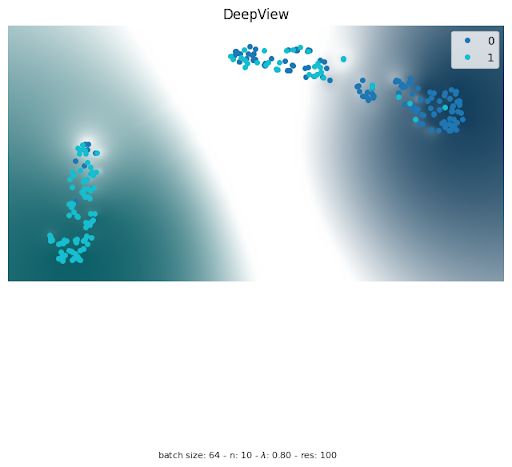}
        \caption{$\lambda=0.8$}
        \label{subfig:lambda0.8}
    \end{subfigure}
    \begin{subfigure}[b]{0.30\textwidth}
        \centering
        \includegraphics[width=\textwidth,trim= 0 150 0 25, clip=true]{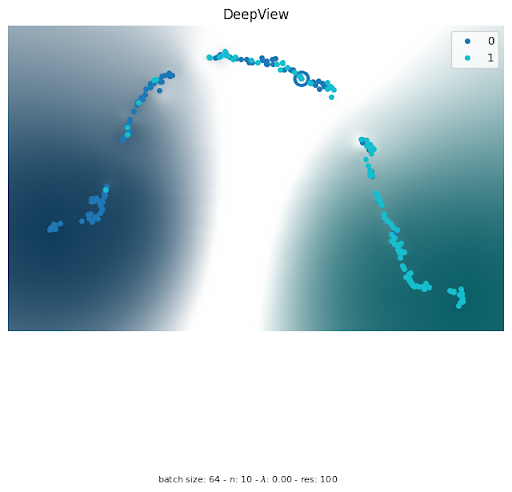}
        \caption{$\lambda=0$}
        \label{subfig:lambda0}
    \end{subfigure}
    \caption{DeepView of the Pre-Trained BERT with classification head trained on SST2, a binary classification dataset; the background colors correspond to either class label; the color intensity is a proxy for classification certainty; one dot corresponds to a single embedded sentence.
    $\lambda$ encodes different amounts of discriminative information.}
    \label{fig2}
\end{figure}

To investigate the effect of discriminative distance in the respective NLP setting, we compare DeepView visualizations based on different amounts of discriminative information, as formalized by $\lambda$ in equation \eqref{eq:dist}. We consider the three different classification model types: PT, FT, and MT BERT for each of the GLUE datasets. Using $Q_{kNN}$ \ref{sec:qknn}, we evaluate for each resulting DeepView visualization whether neighboring data points are classified with the same label by the classifier, i.e.\ the consistency of the displayed class structure. 

More precisely, after having trained the classification models, we select a random sample of 250 data points which were not part of the classification training set and apply DeepView with varying weightings $\lambda$ of the two metrics. Thereby, $\lambda = 1$ refers to using only the cosine distance and the other extreme $\lambda = 0$ to only discriminative distance. 
The average $Q_{kNN}$ values are displayed together with their standard deviations in Table \ref{tab4}. In the fine-tuned and multi-task settings, we averaged over all the datasets, and in the pre-trained case, we averaged over the dataset where the classifier performed better than random.

We can observe two different trends: For the pre-trained models, the error changes heavily as $\lambda$ decreases from $1$. For the two fine-tuned cases, the error values change very little.
Using a larger discriminative weight displays the aspects of the embedding relevant to the downstream task. The different results for pre-trained models show that this information is included in the embedding but not tailored to it. For fine-tuned models, as expected, the representation adjusts to the downstream task, i.e.\ there is less to no difference. We also would like to highlight that DeepView by changing $\lambda$ values allows us to see that!

\begin{figure}[tb]
    \centering
    \begin{subfigure}[b]{0.48\textwidth}
        \includegraphics[width=\textwidth, trim= 20 150 20 170, clip=true]{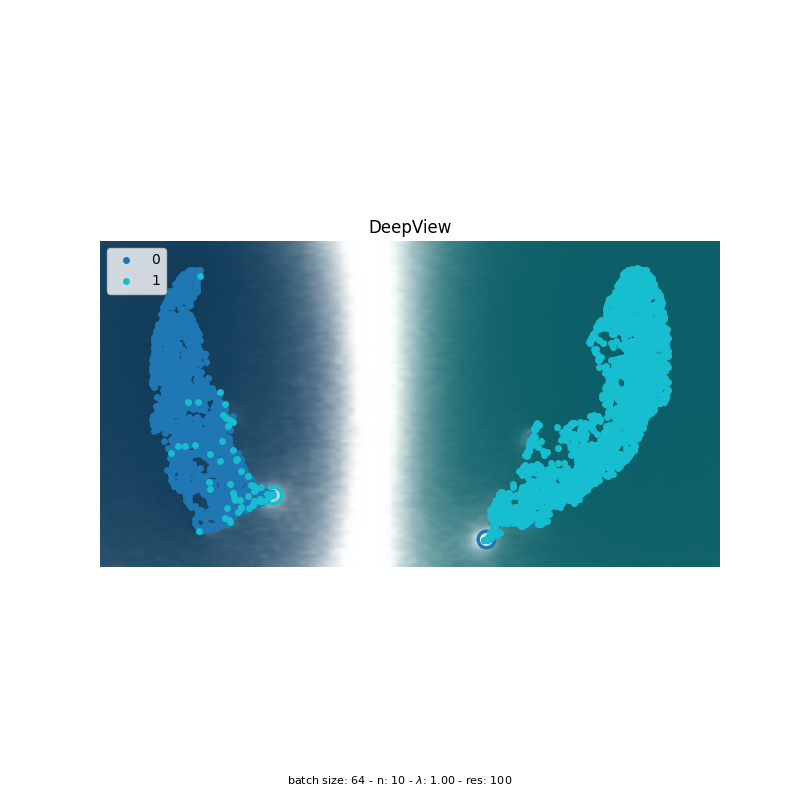}
        \caption{Multi-Task BERT}
        \label{subfig:multitask_bert}
    \end{subfigure}
    \hfill
    \begin{subfigure}[b]{0.50\textwidth}
        \includegraphics[width=\textwidth, trim= 20 150 20 185, clip=true]{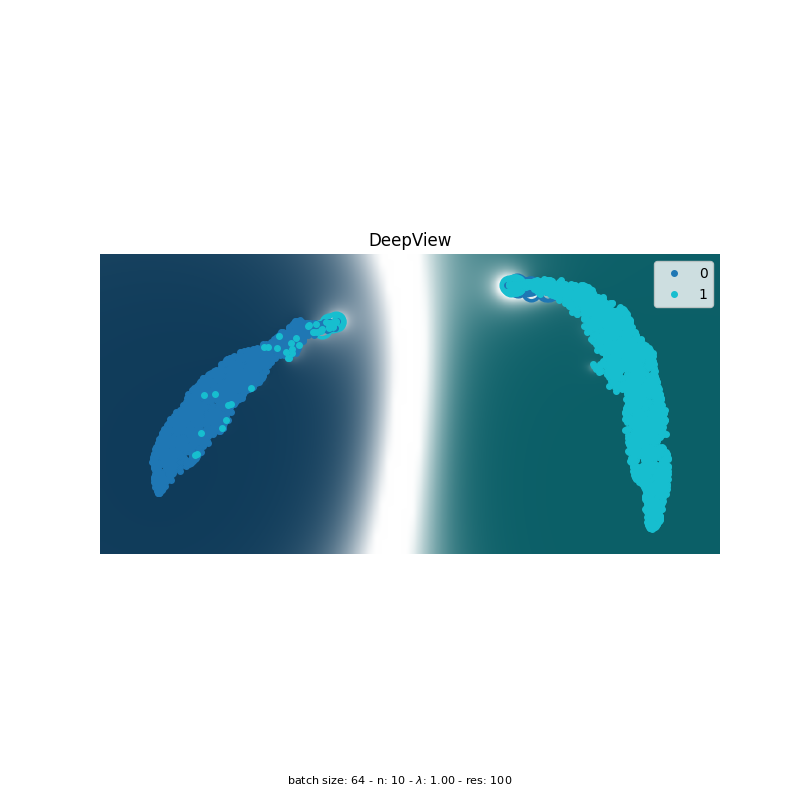}
        \caption{SST2 Fine-tuned BERT}
        \label{subfig:finetuned_bert}
    \end{subfigure}
    \caption{DeepView of the Multi-Task BERT and SST2 Fine-tuned BERT Model's embedding space with respect to the SST2 dataset. Figure \ref{subfig:multitask_bert} also contains an adversarially attacked data point. The bottom left of the cyan region displays an area of high uncertainty which we investigate in  Section \ref{sec:exper_adv}.}
    \label{fig3}
\end{figure}

We also show example visualizations for the SST2 data set. Figure \ref{fig2} shows the DeepView of a pre-trained model for different values of $\lambda$, while Figure \ref{fig3} shows a DeepView for a fine-tuned and a multi-task fine-tuned model, both for $\lambda = 1$. While $\lambda=1$ in Figure \ref{subfig:lambda1} looks like a random embedding regarding the class labels and only shows meaningful class structure as discriminative information is included, in this example with $\lambda=0.8$, in Figure \ref{subfig:lambda0.8} a rather clear class structure is displayed for $\lambda=1$. These results follow the observations in Table \ref{tab4}.
Finally, Figure \ref{subfig:lambda0} with $\lambda=0$ demonstrates why regularizing with an unsupervised distance metric is useful: the fisher metric focuses on class differentiation and thus removes most of the intra-class variation.

Consequently, we will utilize in the following DeepViews $\lambda=1$ if we are visualizing a fine-tuned model, and $\lambda=0.8$ for pre-trained ones, as it corresponds to the largest proportion of unsupervised metric while having a low $Q_{kNN}$ error.

\subsection{Inspecting the Decision Function in the Presence of Adversarially Manipulated Data} \label{sec:exper_adv}
In the following, we demonstrate how DeepView can help us identify malicious data samples in an example containing adversarially manipulated inputs.
 
We utilize samples from the Adversarial Glue dataset curated by \cite{wang2022adversarial}, focusing on a targeted attack method which operates at the sentence level and employs a distraction-based tactic by adding a randomly generated URL to a negative sentiment input. 

As an example, the attack would adapt "I do not like this movie." to e.g.\ "I do not like this movie http://ahsdbw.gos.", which leaves the sentiment of the sentence unchanged. 

For this experiment, we utilize the Multi-Task model and randomly sample 5000 data points from the validation and training set of SST2 and select an adversarial sample at random. Following that, we visualize the decision space using DeepView in Figure \ref{subfig:multitask_bert}. 

Upon closer inspection, we identify particularly difficult areas in the bottom left of the cyan region and bottom right of the dark blue region. The white coloration of the areas indicate difficulty for the model in determining sentiment. In scenarios where adversarial inputs are suspected, thorough investigation of these regions is advisable. The priority is given to the largest uncertain region which we visually determine to be in the cyan region.

\begin{figure}[t]
\centering
\includegraphics[scale=0.20]{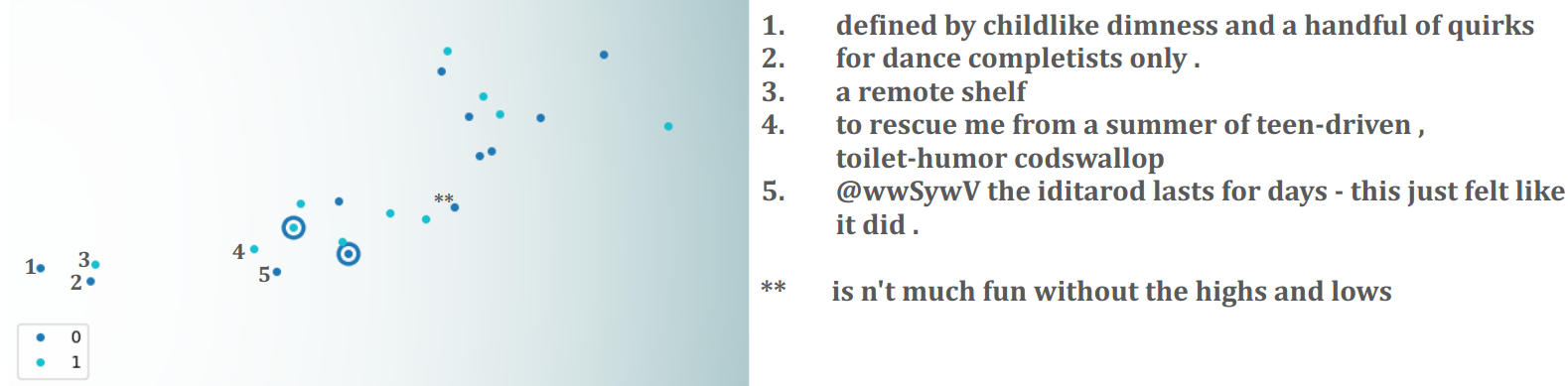}
\caption{Left: This image is zoomed into the uncertain region in the bottom left of the cyan region of Figure \ref{subfig:multitask_bert}. The numbers correspond to the selection order of the points and to the sentences on the right. We selected points from left to right because the background indicates a high level of uncertainty in the left.
The fifth point selected reveals the adversarial attack. 
} \label{adversarial}
\end{figure}

In Figure \ref{adversarial}, we zoom into the scatter plot and observe that the background appears more white to the left of the image than the right, signifying greater uncertainty in the left region. As such, we begin our investigation by selecting the leftmost point, denoted by the number 1 in the image. Upon examination, we find that this is not the perturbed data point. Subsequently, we examine other points and ascertain that the fifth point indeed represents the adversarially perturbed data point.  Additionally, our inspection reveals other noteworthy points located in the region; in particular, the point selected, denoted by asterisks, reveals another type of noise, akin to typo-based attacks \cite{Li_2019}, already present within the SST2 dataset. Thus, DeepView can be used to narrow the search space of potential malicious data points down to isolated areas using the background color as a prioritization guide.

\subsection{Investigating the Embedding Space of the Multi-Task Model} \label{sec:exper_multi} 

\begin{figure}[tb]
\centering
\includegraphics[width=0.4\textwidth,trim= 0 30 0 0, clip=true]{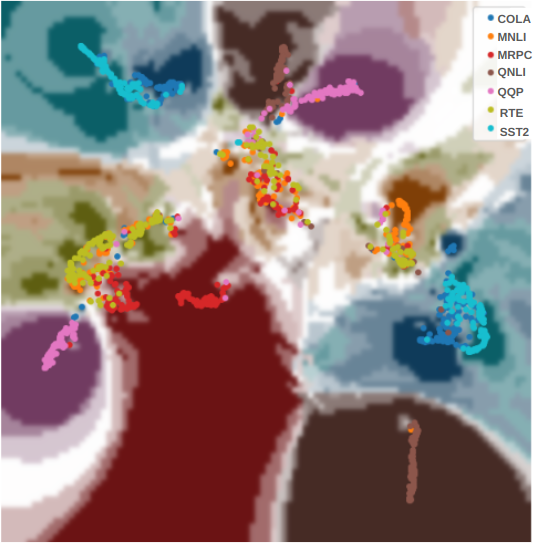} 
\includegraphics[width=0.49\textwidth]{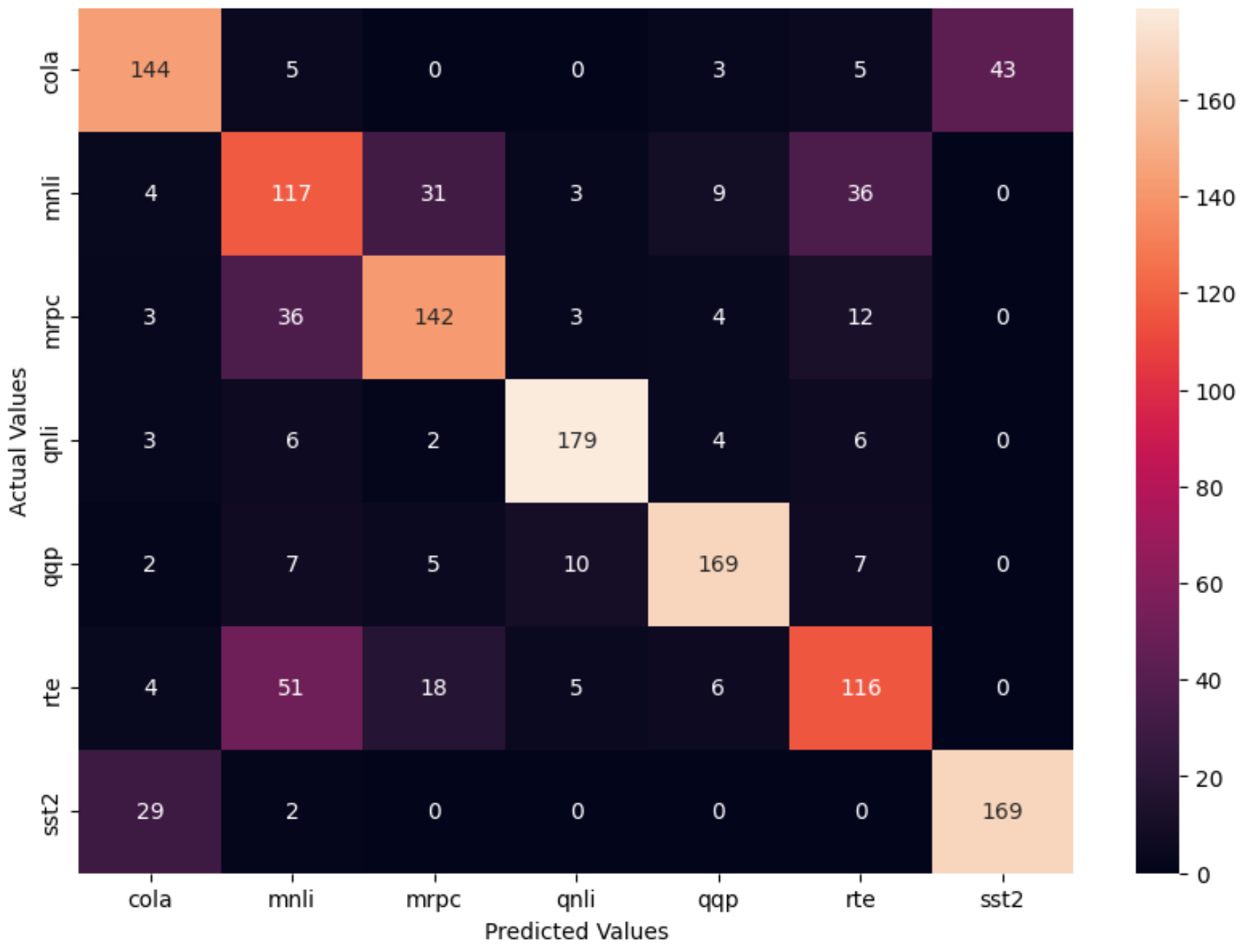}
\caption{Left: DeepView of a kNN classifier on the embedding space of the multi-task model which is trained to differentiate between the datasets. Each point and region are colored according to respective dataset. 
Right: Confusion matrix of the same kNN classifier. }\label{fig5} \label{fig6}
\end{figure}

The GLUE Authors and the broader NLP community agree that RTE is the most challenging task in the GLUE collection \cite{wang2020superglue}. In our studies, we observe a noteworthy improvement between the FT and MT settings (from 0.68 to 0.76), suggesting a training synergy between the datasets. Hence, we utilize DeepView to pinpoint potentially interesting aspects in the embedding space of the MT model and then validate our observations. 
For this purpose, we employ a kNN classifier to differentiate the datasets in the embedding space of the model, with $k=5$. We show the according DeepView in Figure \ref{fig5} (left). 

Initially, we note significant separation among many datasets, despite the MT model being trained with individual classification heads for each dataset i.e. it was not trained for such distinction. Furthermore, when focusing on the RTE dataset, we recognize some possible confusion with MNLI, particularly in the approximated decision function plotted in the background. After DeepView has pointed us to these relations, we verify them with a confusion matrix in Figure \ref{fig6} (right). Indeed, the confusion between RTE and MNLI is the most frequent. This suggests that MT BERT represents these two datasets very similarly. Also, because this model achieves a better accuracy on RTE, it makes the following hypothesis, which we phrase as a question, plausible: Does a representation obtained by fine-tuning on MNLI improve the performance for RTE?

In order to investigate whether a synergy can be obtained, we utilize the MNLI fine-tuned model, remove its classification head, freeze the weights of the encoder and train the weights of a new classification head for RTE. Evaluating the performance, we observe that indeed the accuracy goes up from $0.68$ of the regular FT RTE model to $0.76$ which is comparable to the MT model. We conclude that indeed a synergy has been implemented this way.

\subsection{Local Neighborhood Investigation} \label{sec:exper_neigh}
\begin{figure}[tb]
    \centering
    \begin{subfigure}[b]{0.8\textwidth}
        \includegraphics[width=\textwidth, scale=0.7]{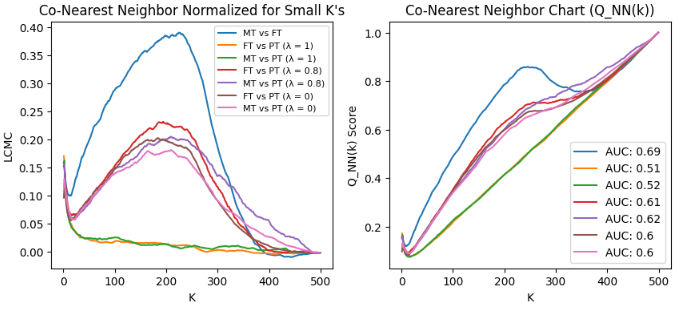}
        \caption{Low Dimensional Neighborhood Analysis}
        \label{subfig:low_dim}
    \end{subfigure}
    \begin{subfigure}[b]{0.8\textwidth}
        \includegraphics[width=\textwidth, scale=0.7]{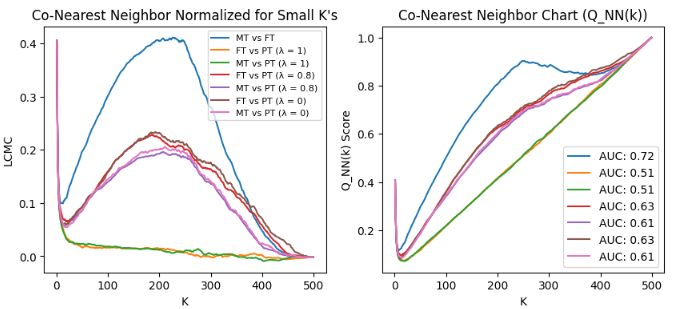}
        \caption{High Dimensional Neighborhood Analysis}
        \label{subfig:high_dim}
    \end{subfigure}
    \caption{$Q_{NN}(k)$ and LCMC curves of models trained on SST2 in high (\ref{subfig:high_dim}) and low (\ref{subfig:low_dim}) dimensions. Visually, we can see each of the curves are very similar between high and low dimensions. The similarity is further proven by the AUC scores located in the bottom right of each figure.}
    \label{neighborhood_high_low}
\end{figure}

In the following, we propose the use of DeepView in a pipeline to compare the classification strategies between two models, allowing us to quantitatively measure their similarity. 
We again use SST2 as a running example.
Here, we apply the scores from \ref{sec:dr_eval} to compare the neighborhood structure of different DeepView embeddings and display the resulting $Q_{NN}$ and LCMC scores in Figure \ref{subfig:low_dim}, as well as the aggregated $Q_{local}$ and $K_{max}$ measures in Table \ref{tab11}.

\begin{table}[tb]
\centering
\caption{SST2 $Q_{NN}(k)$ Statistics. $\lambda$ is displayed in brackets.}\label{tab11}
\begin{tabular}{l ccccc}
\toprule
       & MT vs FT & FT vs PT ($1$) & MT vs PT ($1$) & FT vs PT ($0.8$) & MT vs PT ($0.8$) \\
\midrule
$Q_{local}$ & 0.50     & 0.00                     & 0.08                     & 0.34                       & 0.35  \\ 
$K_{max}$   & 225      & 1                        & 2                        & 194                        & 209 \\ 
\bottomrule
\end{tabular}
\end{table}

\noindent \textbf{Analysis} The largest similarity can be observed between MT and FT in the curves (AUC of $0.69$) as well as with the $Q_{local}$ measure $0.5$. This indicates that the number of co-neighbors in these model settings is the greatest. 
In the pre-trained ($\lambda = 1$) versus multi-task or fine-tuned cases, the $Q_{local}$ score is very low at 0.08 and 0.00, demonstrating that there are very few co-neighbors between the projections which aligns with our results from Section \ref{sec:exper_disc} that no useful representation results without discriminative distances in this case. When incorporating it, as in ($\lambda=0.8$), $Q_{local}$ as well as the AUC scores increase, confirming that with the discriminative distances the representation is more similar to the better performing MT model. The results of the pre-trained models remain consistent across all the models where the pre-trained model outperformed random guessing. In the comparison between fine-tuned and multi-task case, the embedding spaces are consistently more similar in terms of their neighborhoods except in the case of RTE which we investigated in Section \ref{sec:exper_multi}.

Finally, we also evaluate in how much the observations in this section based on the DeepView embeddings resemble the relations in the original space. For this purpose, we rerun the neighborhood comparisons in the high-dimensional space and show them in Figure \ref{subfig:high_dim}. These curves and the AUC scores resemble a high similarity for all cases, indicating that the neighborhoods of each model are represented in both the low-and high- dimensional embedding spaces. Hence, we conclude that it is suitable to examine the neighborhood differences in the representation/visualization created by DeepView.

\section{Conclusions} \label{sec:conclusion}
This study extends the capabilities of DeepView into the domain of NLP. We analyze the embedding space of the popular and widely-used BERT transformer across various text classification tasks using the GLUE dataset in distinct training settings (pre-trained, fine-tuned, and multi-task). 
We demonstrate that DeepView enables us to investigate the downstream task-related part of the embedding space, and for pre-trained models, discriminative distances are crucial for this examination. Further, we show that DeepView enables quick detection of adversarial and atypical data, even among 5000 data points. 
We investigate the embedding of a multi-task model and use DeepView to pinpoint a source of synergy. 
Finally, we also investigate local neighborhoods in the embeddings of DeepView and verify similar relationships between different models.

Promising directions for future work along this line include extending observed phenomena from Pre-Trained BERT to other Language Models,
expanding DeepView to investigate robustness, feature manipulation, and uncertainty. 

\bibliographystyle{splncs04}

\bibliography{bibliography}

\section{Appendix}

\begin{table}
\caption{Pre-Trained $Q_{kNN}$ Errors.}\label{tab9}
\hspace{1.5cm}\begin{tabular}{llllllll}
\toprule
Dataset & \textbf{$\lambda = 0$} & \textbf{$\lambda = 0.2$} & \textbf{$\lambda = 0.4$} & \textbf{$\lambda = 0.6$} & \textbf{$\lambda = 0.8$} & \textbf{$\lambda = 1.0$} \\
\midrule
SST2 & 0 & 0 & 0 & 0 & 0 & 0.368 \\
RTE & 0.004 & 0.004 & 0.004 & 0.004 & 0.004 & 0.032 \\
COLA & 0.016 & 0.016 & 0.02 & 0.016 & 0.012 & 0.012 \\
MRPC & 0 & 0 & 0 & 0 & 0 & 0 \\
QQP & 0 & 0 & 0 & 0 & 0 & 0.28 \\
QNLI & 0.008 & 0.016 & 0.008 & 0.008 & 0.004 & 0.104 \\
MNLI & 0.032 & 0.044 & 0.044 & 0.072 & 0.088 & 0.328 \\
\bottomrule
\end{tabular}
\end{table}

\begin{table}
\caption{Fine-Tuned $Q_{kNN}$ Errors.}\label{tab10}
\hspace{1.5cm}\begin{tabular}{llllllll}
\toprule
Dataset & \textbf{$\lambda = 0$} & \textbf{$\lambda = 0.2$} & \textbf{$\lambda = 0.4$} & \textbf{$\lambda = 0.6$} & \textbf{$\lambda = 0.8$} & \textbf{$\lambda = 1.0$} \\
\midrule
SST2 & 0 & 0 & 0 & 0.004 & 0.004 & 0.02 \\
RTE & 0.008 & 0.02 & 0.008 & 0.02 & 0.016 & 0.036 \\
COLA & 0.008 & 0.008 & 0.008 & 0.008 & 0.008 & 0.008 \\
MRPC & 0.004 & 0.004 & 0.004 & 0.004 & 0.004 & 0.012 \\
QQP & 0.008 & 0.012 & 0.008 & 0.008 & 0.012 & 0.024 \\
QNLI & 0 & 0 & 0 & 0.004 & 0.004 & 0 \\
MNLI & 0.032 & 0.036 & 0.032 & 0.04 & 0.048 & 0.044 \\
\bottomrule
\end{tabular}
\end{table}

\begin{table}
\caption{Multi-Task $Q_{kNN}$ Errors.}\label{tab11}
\hspace{1.5cm}\begin{tabular}{llllllll}
\toprule
Dataset & \textbf{$\lambda = 0$} & \textbf{$\lambda = 0.2$} & \textbf{$\lambda = 0.4$} & \textbf{$\lambda = 0.6$} & \textbf{$\lambda = 0.8$} & \textbf{$\lambda = 1.0$} \\
\midrule
SST2 & 0.004 & 0.004 & 0.004 & 0.004 & 0.008 & 0.008 \\
RTE & 0.008 & 0.004 & 0.004 & 0.016 & 0.016 & 0.012 \\
COLA & 0.004 & 0.004 & 0.004 & 0.004 & 0.004 & 0.008 \\
MRPC & 0.004 & 0.008 & 0.008 & 0.012 & 0.012 & 0.016 \\
QQP & 0.02 & 0.012 & 0.008 & 0.008 & 0.008 & 0.008 \\
QNLI & 0 & 0 & 0 & 0.004 & 0.004 & 0.004 \\
MNLI & 0.012 & 0 & 0.008 & 0 & 0.008 & 0.008 \\
\bottomrule
\end{tabular}
\end{table}

\begin{table}
\caption{Pre-Trained $Q_{data}$ Errors.}\label{tab7}
\hspace{1.5cm}\begin{tabular}{llllllll}
\toprule
Dataset & \textbf{$\lambda = 0$} & \textbf{$\lambda = 0.2$} & \textbf{$\lambda = 0.4$} & \textbf{$\lambda = 0.6$} & \textbf{$\lambda = 0.8$} & \textbf{$\lambda = 1.0$} \\
\midrule
SST2 & 0 & 0 & 0 & 0 & 0 & 0.4 \\
RTE & 0 & 0 & 0 & 0.013 & 0.01333 & 0.013\\
COLA & 0 & 0 & 0 & 0.013 & 0.027 & 0.133 \\
MRPC & 0 & 0 & 0 & 0 & 0 & 0 \\
QQP & 0 & 0 & 0 & 0 & 0 & 0.387 \\
QNLI & 0 & 0 & 0.0267 & 0.013 & 0 & 0.107 \\
MNLI & 0.027 & 0.013 & 0.067 & 0.093 & 0.12 & 0.36 \\
\bottomrule
\end{tabular}
\end{table}

\begin{table}
\caption{Fine-Tuned $Q_{data}$ Errors.}\label{tab6}
\hspace{1.5cm}\begin{tabular}{llllllll}
\toprule
Dataset & \textbf{$\lambda = 0$} & \textbf{$\lambda = 0.2$} & \textbf{$\lambda = 0.4$} & \textbf{$\lambda = 0.6$} & \textbf{$\lambda = 0.8$} & \textbf{$\lambda = 1.0$} \\
\midrule
SST2 & 0 & 0 & 0 & 0 & 0 & 0 \\
RTE & 0.027 & 0 & 0 & 0 & 0.013 & 0.013 \\
COLA & 0 & 0 & 0 & 0 & 0 & 0 \\
MRPC & 0 & 0 & 0 & 0 & 0 & 0.013 \\
QQP & 0 & 0 & 0 & 0.013 & 0.013 & 0.027 \\
QNLI & 0 & 0 & 0 & 0.013 & 0.013 & 0 \\
MNLI & 0 & 0.027 & 0.027 & 0.013 & 0.013 & 0.013 \\
\bottomrule
\end{tabular}
\end{table}

\begin{table}
\caption{Multi-Task $Q_{data}$ Errors.}\label{tab8}
\hspace{1.5cm}\begin{tabular}{llllllll}
\toprule
Dataset & \textbf{$\lambda = 0$} & \textbf{$\lambda = 0.2$} & \textbf{$\lambda = 0.4$} & \textbf{$\lambda = 0.6$} & \textbf{$\lambda = 0.8$} & \textbf{$\lambda = 1.0$} \\
\midrule
SST2 & 0 & 0 & 0.013 & 0 & 0.013 & 0.013 \\
RTE & 0 & 0.027 & 0.013 & 0.013 & 0 & 0.027 \\
COLA & 0 & 0 & 0 & 0 & 0 & 0.013 \\
MRPC & 0 & 0 & 0 & 0 & 0 & 0 \\
QQP & 0 & 0 & 0 & 0.013 & 0.013 & 0.013 \\
QNLI & 0 & 0 & 0.013 & 0.013 & 0.013 & 0.013 \\
MNLI & 0 & 0 & 0 & 0 & 0 & 0 \\
\bottomrule
\end{tabular}
\end{table}

\end{document}